%% file: emnlp2021.tex
\pdfoutput=1
\documentclass[11pt]{article}

\usepackage{emnlp2021}

\usepackage{times}
\usepackage{latexsym}

\usepackage[T1]{fontenc}
\usepackage[utf8]{inputenc}

\usepackage{microtype}

\usepackage{amsmath}
\usepackage{amssymb}
\usepackage{amsthm}
\usepackage{amscd}
\usepackage{bm}
\usepackage{graphicx}

\usepackage{booktabs}
\usepackage{xspace}
\usepackage{tikz}
\usepackage{xcolor}
\usepackage{graphics}
\usepackage{multirow}
\usepackage{multicol}
\usepackage[super]{nth}
\usepackage{lipsum}
\usepackage{enumitem} % for noindent itemize

\usepackage{scalefnt} % for scaling digits in textsc
\usepackage{xfrac} % lazy fractions \sfrac

\newcommand{\minus}{%
\parbox{0.15cm}{\centering\tikz{\draw[-](0cm,0)--(0.15cm,0);}}
}

\newcommand\smaller[2][0.85]{{\scalefont{#1}#2}}

\newcommand{\Tab}[1]{Tab.~\ref{#1}}
\newcommand{\Fig}[1]{Fig.~\ref{#1}}

\newcommand{\Equ}[1]{Eq.~\eqref{#1}}
\newcommand{\Sec}[1]{Section~\ref{#1}}

\newcommand{\xT}{\ensuremath{x_{\textsc{t}}}\xspace}
\newcommand{\xG}{\ensuremath{x_{\textsc{g}}}\xspace}

\newcommand{\hxG}{\ensuremath{\hat{x}_{\textsc{g}}}\xspace}
\newcommand{\hxT}{\ensuremath{\hat{x}_{\textsc{t}}}\xspace}

\newcommand{\sxT}{\ensuremath{x^{*}_{\textsc{t}}}\xspace}

\newcommand{\gtt}{\ensuremath{\textsc{g\smaller{2}t}}\xspace}  % g2t
\newcommand{\ttg}{\ensuremath{\textsc{t\smaller{2}g}}\xspace}  % t2g

\newcommand{\hw}{\ensuremath{\hat{w}}\xspace}  % w^
\newcommand{\Lce}{\ensuremath{\mathcal{L}_{\textsc{ce}}}\xspace}
\newcommand{\Lrl}{\ensuremath{\mathcal{L}_{\textsc{rl}}}\xspace}
\newcommand{\Lscst}{\ensuremath{\mathcal{L}_{\textsc{scst}}}\xspace}

\newcommand{\mT}{\mathcal{M}_{\textsc{t}}\xspace}
\newcommand{\mG}{\mathcal{M}_{\textsc{g}}\xspace}
\newcommand{\mTG}{\mathcal{M}_{\textsc{t+g}}\xspace}

\newcommand\sbullet[1][.5]{\mathbin{\vcenter{\hbox{\scalebox{#1}{$\bullet$}}}}}

\newcommand{\tekgen}{\textsc{TekGen}\xspace}

\title{ReGen: Reinforcement Learning for Text and Knowledge Base Generation using Pretrained Language Models}

\author{Pierre L. Dognin  \\ IBM Research \\ \texttt{pdognin@us.ibm.com} \And
  Inkit Padhi \\ IBM Research \\ \texttt{inkpad@ibm.com} \\ \AND
  Igor Melnyk \\ IBM Research \\ \texttt{igor.melnyk@ibm.com} \And
  Payel Das\\IBM Research \\ \texttt{daspa@us.ibm.com}}
\begin{document}
\maketitle

% Automatic construction of relevant Knowledge Bases (KBs) from text, and generation of semanticly meaningful text from KBs are both long-standing goals in Machine Learning.
% In this paper, we present ReGen, a bidirectional generation of text and graph leveraging Reinforcement Learning to improve the performance. 

%%% DO NOT CHANGE THE FIRST 2 SENTENCES IN THE ABSTRACT. 
%%% WE USE THEM AS INPUT FOR GENERATION EXAMPLES IN FIG1 !!
\begin{abstract}
Automatic construction of relevant Knowledge Bases (KBs) from text, and generation of semantically meaningful text from KBs are both long-standing goals in Machine Learning.
In this paper, we present ReGen, a bidirectional generation of text and graph leveraging Reinforcement Learning (RL) to improve performance. 
% DO NOT CHANGE ABOVE
Graph linearization enables us to re-frame both tasks as a sequence to sequence generation problem regardless of the generative direction, which in turn allows the use of Reinforcement Learning for sequence training where the model itself  is employed as its own critic leading to Self-Critical Sequence Training (SCST). 
We present an extensive investigation demonstrating that the use of RL via SCST benefits graph and text generation on WebNLG+ 2020 and \tekgen datasets.
Our system provides state-of-the-art results on WebNLG+ 2020 by significantly improving upon published results from the WebNLG 2020+ Challenge for both text-to-graph and graph-to-text generation tasks.
\end{abstract}

\input{intro.tex}
\input{related_work.tex}
\input{models.tex}

\input{experiments.tex}

\input{conclusions.tex}

% \section*{Acknowledgements}

\bibliography{anthology,emnlp2021}
\bibliographystyle{acl_natbib}

\clearpage

\appendix
\input{suppl.tex}

\end{document}

%% file: intro.tex
\section{Introduction}

Graph representation of knowledge is a powerful tool to capture real-world information where complex relationships between node entities can be simply encoded.
Automatic generation of Knowledge Bases (KBs) from free-form text and its counterpart of generating semantically relevant text from KBs are both active and challenging research topics.

Recently, there has been an increased interest in leveraging Pretrained Language Models (PLMs) to improve performance for text generation from graph, or graph-to-text (G2T) task \citep{ribeiro2020investigating}.
Indeed, large PLMs like T5 \citep{raffel2020exploring} and BART \citep{lewis-etal-2020-bart} that have been pretrained on vast amount of diverse and variedly structured data, are particularly good candidates for generating natural looking text from graph data. 

BART- and T5-related models have been employed by top performers in public challenges such as the WebNLG+ 2020 Challenge \citep{webnlg-2020-international} where both graph-to-text and text-to-graph (T2G) tasks are offered, under the names \emph{RDF-to-Text} and \emph{Text-to-RDF} (semantic parsing) respectively (RDF stands for Resource Description Framework, a standard for describing web resources). 
One can notice that  more teams entered the competition for the G2T task than for T2G as the latter is a much harder task.
Best models generally use PLMs and fine-tune them for the target modality at hand (either graph or text).
This is possible by re-framing the T2G and G2T generations as a sequence to sequence (Seq2Seq) generation problem, which suits fine-tuning PLMs well.
One can therefore hope to leverage the large pretraining of PLMs to improve the overall quality of generation.

The Seq2Seq formulation requires a linearization of any input graph, which is not unique. This creates an opportunity for data augmentation where multiple linearizations are provided to the model at training time so the model learns the content represented by the graph, not the order of its sequential representation.

\begin{figure*}[ht!]
\includegraphics[width=\textwidth]{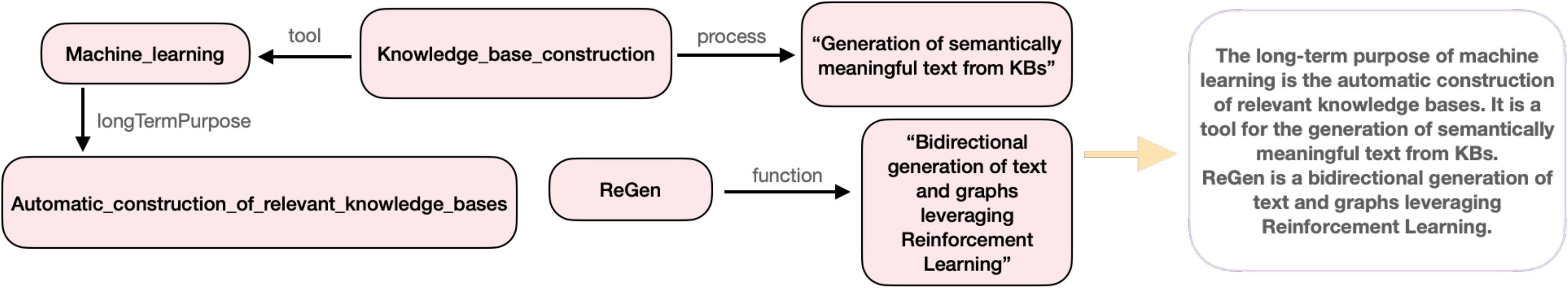}
\caption{Actual examples of generation for Text-to-Graph and Graph-to-Text tasks using our best RL models. The first two sentences of the abstract were processed through our best models. First, a graph was created capturing the facts from the input sentences. Then, this graph was used as input to generate text. Despite a strong domain mismatch between input data and models, the generated paragraph is capturing most of the original sentences content. Both models were trained using RL, specifically Self-Critical Sequence Training (SCST).}
\label{fig:gen}
\end{figure*}

In this work, we are interested in leveraging the power of PLMs for both G2T and T2G generation tasks, and will demonstrate the strength of our approach by improving upon the best results of the WebNLG+ 2020 Challenge (rev 3.0) as reported by \citet{castro-ferreira-etal-2020-2020} for both T2G (Semantic Parsing) and G2T (Data-to-Text) tasks.
We will also present results for the \tekgen Corpus \citep{agarwal2021knowledge} to show performance on a different, much larger dataset.
To illustrate the task of generation, \Fig{fig:gen} provides examples of G2T and T2G outputs obtained using the proposed generation framework. The first two sentences of the abstract of this paper were used as input for T2G using our best model. The model generates a graph from the input text by simultaneously extracting relevant nodes and linking them coherently. 
For the G2T task, another model starts from the generated graph and generates semantically relevant text from it. 
As one can appreciate, the final text is quite readable and captures most facts from the original abstract sentences despite a strong domain  mismatch between input data and training data, which both models were built on.

Since both T2G and G2T generative tasks can be formulated as a Seq2Seq problem, we propose to use Reinforcement Learning (RL) as part of the PLMs fine-tuning on the target domain data. 
For both G2T and T2G tasks, a differentiable function such as the cross-entropy (CE) loss function is often used, since minimizing it results in maximizing the probability of generating the correct token/word.
However, when it comes to evaluating a model's performance, benchmarks often use BLEU \citep{pa-pa-aung-etal-2020-automatic}, METEOR \citep{lavie-agarwal-2007-meteor}, chrF++ \citep{popovic-2017-chrf} for G2T, or simply F1, Precision, and Recall scores for T2G, none of which being differentiable.
During training, one hopes that by minimizing the CE loss, the model will tend towards better prediction of the target tokens, hence improving on evaluation metrics as a beneficial by-product. 
Thankfully, RL provides a framework where we can update our model parameters so to improve evaluation metrics directly. 
Mixed Incremental Cross-Entropy Reinforce (MIXER) from \citet{ranzato2016sequence} introduced using REINFORCE \citep{williams1992simple} for sequence training. 
We propose to use one of its variant known as Self-Critical Sequence Training (SCST) \citep{rennie2017self} for both T2G and G2T training.

To summarize, our main contributions are:

\noindent$\sbullet[.70]$ We propose to use RL-based sequence training, specifically SCST, for both G2T and T2G tasks. This is the first time that RL based training is proposed to the bi-directional generation of text and graph.
To the best of our knowledge, the present work is the first time it is introduced for a T2G task.

\noindent$\sbullet[.70]$ We demonstrate that our approach provides better performance than the best systems reported for the  WebNLG 2020+ Challenge.

\noindent$\sbullet[.70]$ We provide a thorough investigation of SCST-based training for both T2G and G2T tasks, including best rewards combination.

\noindent$\sbullet[.70]$ We constructed subject and relation-object boundaries from \tekgen sentence-triples pairs and showed performance of our approach for both T2G and G2T tasks. 

\noindent$\sbullet[.70]$ We adapted the large-scale \tekgen corpus \citep{agarwal2021knowledge} for T2G and G2T tasks and confirmed the benefit of SCST-based fine-tuning approach over CE-trained baselines. 

%% file: related_work.tex
\section{Related work}
\label{sec:related}
In the  WebNLG+ 2020 Challenge, most top performing models relied on  fine-tuning of PLMs. 
Interestingly, all four top teams in the Challenge proposed quite different approaches while leveraging PLMs. 
\nth{1} place Amazon AI \citep{guo-etal-2020-2} pipelined a relational graph convolutional network (R-GCN) and a T5 PLM with some canonicalization rules. \nth{2} place OSU Neural NLG \citep{li-etal-2020-leveraging}, the closest to our approach in spirit, used T5 and mBART PLMs to fine-tune after special data preprocessing. \nth{3} place FBConvAI  \citep{yang-etal-2020-improving} used BART PLM and multiple strategies to model input RDFs. \nth{4} place bt5 employed a T5 PLM trained in a bi-lingual approach on English and Russian, even using WMT English/Russian parallel corpus.

Recently, \citet{dognin-etal-2020-dualtkb, guo-etal-2020-cyclegt} proposed models trained to generate in both T2G and G2T directions, with consistency cycles created to enable the use of unsupervised datasets. In contrast, our approach of fine-tuning a T5 PLM is fully supervised but can produce either the specialized models for T2G and G2T tasks alone, or a hybrid model that can handle both T/G inputs simultaneously to generate the corresponding translated outputs G/T.

Note that in contrast to many WebNLG+ 2020 Challenge participants, e.g. \cite{li-etal-2020-leveraging}, no pre-processing of the data is performed for text, while for graph triples, we add tokens to mark subject, predicate, and object positions in their linearized sequence representation. Moreover, data augmentation is performed by allowing random shuffling of triples order in graph linearization to avoid a model to learn the exact order of triples, especially for the T2G task. 

While the use of RL training in PLM has been explored in many works, the approach of \citep{Chen2020Reinforcement} is closest to ours. However, their work focuses on the improved text generation in the context of natural question generation, while in our algorithm we use it for graph-to-text and text-to-graph generations.

%% file: models.tex
\section{Models}

Models are trained on a dataset $\mathcal{D}$ composed of a set of $(\xT, \xG)^i$ samples, where $\xT$ is made of text (one or more sentences), and $\xG$ is a graph represented as a list of triples $\xG = [(s^1,p^1,o^1),\dots,(s^K, p^K, o^K)]$, where the $k$-th triple is composed of a subject $s^k$, predicate (relationship) $p^k$, and object $o^k$. The superscript $i$ denotes the $i$-th sample in $\mathcal{D}$.
For G2T, the model is given $\xG$ as input and must generate $\hxT$.
A cross-entropy loss is computed as an expectation: 
\begin{align}
       \Lce^{\textsc{t}} = \underset{\xT \sim \mathcal{D}}{\mathbb{E}} \left[\minus \log p_{\theta}^{\gtt}(\xT)\right],
        \label{eq:CE1}
\end{align}
where $p_{\theta}^{\gtt}(\xT)$ is the distribution of the generated sequence $\hxT = T_{\gtt}(\xG)$, $T_{\gtt}(\xG)$ being the transformation from graph to text, and where our model is parameterized by $\theta$. 
$\hxT = [\hw_1,\hw_2,\dots,\hw_T]$ is a sequence of generated tokens/words.
Similarly, for training a T2G model, the cross-entropy loss used in training is simply
\begin{align}
        \Lce^{\textsc{g}} = \underset{\xG \sim \mathcal{D}}{\mathbb{E}} \left[\minus \log p_{\theta}^{\textsc{t\smaller{2}g}}(\xG)\right],
        \label{eq:CE2}
\end{align}
where $p_{\theta}^{\ttg}(\xG)$ is the distribution of the generated graph $\hxG = T_{\ttg}(\xT)$, $T_{\ttg}(\xT)$ being the transformation from text to graph. 

In both \Equ{eq:CE1} and \Equ{eq:CE2}, $\xG$ must be expressed as a sequence of tokens $t_j$
such that a list of triples $\xG$ turns into a list of tokens $[t_1, t_2,\cdots, t_M]$. 
This is simply done by adding tokens marking the subject, predicate, and object locations in the sequence such that each triple $(s^k,p^k,o^k)$ is turned into a sequence such as $[\textsc{<s>}, w_1^s, \textsc{<p>}, w_1^p, w_2^p, \textsc{<o>}, w_1^o,w_2^o,w_3^o]$, assuming our subject is made of 1 token, our predicate of 2 tokens, and our object of 3 tokens in this example. $\textsc{<s>,<p>}$, and $\textsc{<o>}$ are just special marker tokens to help the model know where subject, predicate and objects are located in the sequence.

\begin{figure*}[th!]
\centering
\includegraphics[width=\textwidth]{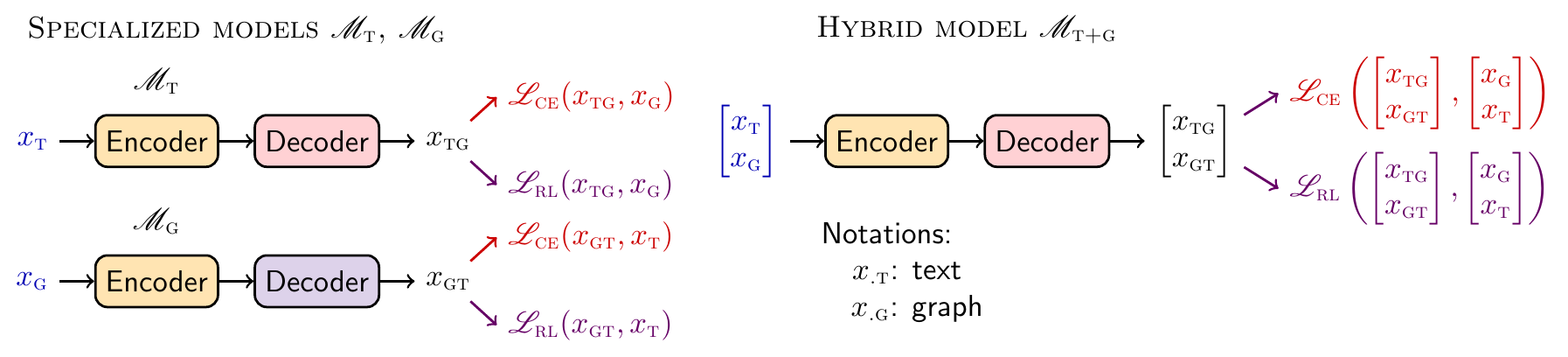}
\caption{Specialized and hybrid models rely on the same losses for fine-tuning. However, specialized models are dedicated to a particular generation task while hybrid models can handle both generation directions.}
\label{fig:models}
\end{figure*}

We start from a pretrained encoder-decoder $\mathcal{M}$ model that we fine-tune on either T2G to get $\mT$, or G2T task to get $\mG$. 
We also propose a third kind of model $\mTG$ to be fine-tuned on \emph{both} T2G and G2T samples, i.e. the model will learn to generate in any direction, by supplying an input sample $x = [ \xT; \xG ]^{\top}$ and corresponding target for it. Input from each modality is prefixed by a task specific string to distinguish samples ("Text to Graph:" for $\xT$ and "Graph to Text:" for $\xG$). 
For $\mTG$ models, the cross-entropy loss is similarly defined as for \Equ{eq:CE1} and \Equ{eq:CE2} such that $\Lce^{\textsc{t+g}} = \underset{x \sim \mathcal{D}}{\mathbb{E}} \left[\minus \log p_{\theta}(x)\right]$. All models are shown in \Fig{fig:models}.
By convention, we refer to models in this paper by their input modality $\textsc{t}$, $\textsc{g}$, or $\textsc{t+g}$.

%%%%%%% MODELS FIGURE

\subsection{Reinforcement Learning}

A sequence generation task can be re-framed as a model picking the best word within a vocabulary to react to its environment and accounting for past predictions,  we can then reformulate Seq2Seq generation into the Reinforcement Learning framework.
A model is an agent that defines a policy resulting in the action of selecting each word during generation, as first introduced by \citet{ranzato2016sequence}.
REINFORCE,  presented by \citet{williams1992simple}, allows the optimization of a model's parameters $\theta$ by maximizing the expected value of the reward $R(\hxT)$ of generated sequence $\hxT = [\hw_1,\dots,\hw_T]$.
To match usual Deep Learning conventions, we can minimize a loss expressed as its negative value:
\begin{align}
    \Lrl &= \minus \! \sum_{\hw_1,\dots,\hw_T} \! p_{\theta}(\hw_1,\dots,\hw_T) r(\hw_1,\dots,\hw_T) \nonumber \\
    &= \minus \, \mathbb{E}_{[\hw_1,\dots,\hw_T]\sim p_\theta} R(\hw_1,\dots,\hw_T), \nonumber \\
    &= \minus \, \mathbb{E}_{\hxT \sim p_\theta} R(\hxT). 
\end{align}
$R(\hxT)$ is the reward for the generated text which is often associated with non-differentiable metrics such as BLEU, METEOR, chrF, etc. We circumvent this issue by using the REINFORCE policy gradient method:
\begin{align}
    \nabla_{\theta} \Lrl \propto \minus \left(R(\hxT) \, \minus \, b\right) \nabla_{\theta} \log p_{\theta}(\hxT), 
\end{align}
where $b$ is a baseline used to reduce the variance of our gradient estimate. 
$b$ can be any function, even a random variable, as long as it is independent of the actions taken to generate $\hxT$, as described in Chapter 13.4 from \citet{suttonRL}.
In Self-Critical Sequence Training (SCST) \citep{rennie2017self}, $b$ is chosen to be the reward of $\sxT$, the output generated by the model by greedy max generation, hence the model serving as its own critic:
\begin{align}
    \nabla_{\theta} \Lscst\! \propto \! \minus \left(R(\hxT)\minus R(\sxT)\right) \nabla_{\theta} \log p_{\theta}(\hxT), 
\end{align}
where $\hxT$ is sampled from our model and $\sxT$ is generated by greedy max.
An interesting property of the baseline is that if $R(\hxT) > R(\sxT)$, sampled $\hxT$ has higher reward than $\sxT$, then the model is updated to reinforce the choices made by this generation. 
In the opposite case where $R(\hxT) < R(\sxT)$, the model update will take the negative gradient or the opposite model update to subdue such generation. 
When $R(\hxT)= R(\sxT)$, no update is performed on  the model since the gradient is effectively zeroed out, regardless of the individual values $R(\hxT)$ and $R(\sxT)$.
This happens when $\hxT$ and $\sxT$ are identical (greedy-max and sampled sequences are identical). 
In that case the sample is lost for RL as no update to the model will result from this sample. 
Basically, REINFORCE is a Monte Carlo method of learning where a gradient update is applied in the direction decided by how $R(\hxT)$ compares to baseline $b$, the role of $b$ being to reduce the variance of the gradient estimate. 
Variations around REINFORCE exist on how to apply the gradients, such as MIXER \citep{ranzato2016sequence}, or on how to evaluate the baseline \citep{luo2020better} to minimize the gradient variance.

In our training, PLMs are first fine-tuned using $\Lce$ loss. Once they reach a good generation quality, the training is switched to RL fine-tuning by minimizing $\Lscst$.

%% file: experiments.tex
\section{Experimental Setup}

In this Section, we present the experimental setup used for all the results reported in this paper.

\noindent\textbf{Models} We used T5 PLMs (from \citet{wolf-etal-2020-transformers}) for our experiments for two distinct models, \emph{t5-large} [770M parameters] and \emph{t5-base} [220M parameters], with a special focus on t5-large as it is the best performing of the two on various NLP tasks.
Models were fine-tuned to be either specialized on T2G ($\mT$) or G2T ($\mG$) task, or to accommodate both directions of generation ($\mTG$). 

\begin{table*}[ht!]
\centering
\begin{tabular}{lcccccc}
%\toprule
WebNLG G2T                & BLEU$\uparrow$ & \multirow{2}{2cm}{\centering BLEU$\uparrow$ NLTK} & METEOR$\uparrow$ & chrF++$\uparrow$  \\
Team/model                    &      &      &           &         \\
\midrule
Amazon AI (Shanghai) \citep{guo-etal-2020-2}          & 0.540 & 0.535 & 0.417 & 0.690 \\
OSU Neural NLG \citep{li-etal-2020-leveraging}        & 0.535 & 0.532 & 0.414 & 0.688 \\
FBConvAI \citep{yang-etal-2020-improving}             & 0.527 & 0.523 & 0.413 & 0.686 \\ 
bt5 \citep{agarwal-etal-2020-machine}                 & 0.517 & 0.517 & 0.411 & 0.679 \\
\midrule
ReGen (Ours) G2T.CE t5-large    &  0.553 & 0.549 &  0.418 &  0.694 \\       
ReGen (Ours) G2T.RL t5-large    &  \textbf{0.563} & \textbf{0.559} &  \textbf{0.425} &  \textbf{0.706}  \\       
\midrule
ReGen (Ours) G2T.CE.ES t5-base (early CE)    &   0.522 & 0.518 &  0.404 &  0.675 \\       
ReGen (Ours) G2T.RL.ES t5-base (early CE)    &   0.531 & 0.527 &  0.410 &  0.686 \\       
\midrule
ReGen (Ours) G2T.CE.best t5-base (best CE)   &  0.524 & 0.520 &  0.404 &  0.677 \\       
ReGen (Ours) G2T.RL.best t5-base (best CE)   &  0.527 & 0.523 &  0.408 &  0.681 \\
\bottomrule
\end{tabular}
\caption{G2T Best results on WebNLG 2020 Challenge (v3.0) dataset. The first four rows were the top performers of the Challenge. Results for CE and RL models are presented for our ReGen systems so to show gains from using SCST. Our G2T.RL is the best system overall, fine-tuning a t5-large model using METEOR reward. G2T.RL.ES and G2T.RL.best show the impact of using early stopping (ES) or best CE selection for starting SCST fine-tuning on a t5-base smaller model while using \textsc{BLEU\_NLTK} reward.}
\label{tab:web_g2t_post}
\end{table*}

\noindent\textbf{Data processing} Graphs are often represented as list of triples. However our model expects a sequence of input words/tokens to work on. 
The linearization of graph triples is obviously ambiguous as there are many ways to traverse a graph (Breadth First Search, Depth First Search, random walk, etc.). 
In practice, we linearize the triples in the order of the list provided by the dataset, but use this inherent linearization ambiguity as an opportunity to do data-augmentation.
Indeed, models are first fine-tuned using cross-entropy loss that strongly penalizes generation if it is in any different order than the ground truth order.
To avoid the model to overfit to our data and memorize observed triples order, we augment the data by including a few permutations of the graph triples.

During graph linearization, we encode the subject, predicate, and object positions by using $\textsc{<s>,<p>,<o>}$ tokens. In practice, we expand the model vocabulary with these special indivisible tokens that are not split during tokenization. No other preprocessing is done on the data for training.
We explored masked and span-masked LM fine-tuning to match T5 pretraining \citep{raffel2020exploring} which did not lead to any noticeable improvements.

\subsection{Datasets}
\label{subsec:datasets}
\noindent\textbf{WebNLG+ 2020} We report results on WebNLG+ 2020 (v3.0) used in the WebNLG 2020 Challenge \citep{webnlg-2020-international}. 
The Challenge comprises of two tasks: RDF-to-text generation (G2T), and Text-to-RDF semantic parsing (T2G). 
The Resource Description Framework (RDF) language is used to encode DBpedia and is commonly used in linked data framework. WebNLG+ uses RDF to encode graphs as sets of triples which are associated to one or more lexicalizations of one or more sentences each. 
Data for English and Russian are provided, but we only worked on the English subset made of 13,211 train, 1,667 dev, 2,155 testA (semantic parsing), and 1,779 testB (data-to-text) samples (triples sets w/ lexicalizations). 
The data is clustered semantically into 16 categories \emph{seen} in train and dev sets (Airport, Astronaut, Building, etc.), while 3 categories (Film, Scientist, and Musical-Work) were introduced in test and are \emph{unseen}, i.e. not present in training; see \citet{castro-ferreira-etal-2020-2020} for more details. 
Results are aggregated for \emph{all}, \emph{seen}, and \emph{unseen} categories during evaluation. Note that in the literature, prior work sometimes report `WebNLG' results on previous dataset version, with completely different performance ranges. 
We compare all our results to WebNLG+ 2020 (v3.0) numbers reported by \citet{castro-ferreira-etal-2020-2020} in their Table 6 for G2T, and Table 10 for T2G tasks, using the provided official scoring scripts.

\noindent\textbf{\tekgen} To further study the robustness of our system, we also provide experiments using \tekgen dataset recently introduced in \citet{agarwal2021knowledge}. The graph-sentence alignments are curated using Wikipedia and Wikidata. This serves as a perfect large scale test-bed for both G2T and T2G tasks. Unfortunately, this dataset lacks in entity/relation/object boundaries, which makes it difficult to evaluate systems for T2G tasks. In order to address this issue, we further process the triple-text (with no triple boundaries) to create list of triples using Wikidata properties lookup, via Wikidata Query Service. Additionally, we limit the validation set and test set to 5K and 50K sentence-triples pairs respectively. Our training split after processing contains 6.3 million sentence-triples pairs.  As a contribution to the work, we will open-source this further processed \tekgen dataset with appropriate subject, object and relation boundaries, which enables conventional evaluation of research systems. An example of the processed \tekgen is shown in \Fig{fig:tekgen_example} in Appendix.

\begin{table*}[ht!]
\centering
\begin{tabular}{llccc}
%\toprule
WebNLG T2G               & Match & F1$\uparrow$ & Precision$\uparrow$ & Recall$\uparrow$  \\
Team/model        &       &              &                      &        \\
\midrule
\multirow{4}{*}{Amazon AI (Shanghai) \citep{guo-etal-2020-2}}  
                        & Exact     &  0.689 &     0.689 &  0.690 \\
                        & Ent\_Type &  0.700 &     0.699 &  0.701 \\
                        & Partial   &  0.696 &     0.696 &  0.698 \\
                        & Strict    &  0.686 &     0.686 &  0.687 \\
\midrule
\multirow{4}{*}{bt5 \citep{agarwal-etal-2020-machine} }
                        & Exact     &  0.682 &     0.670 &  0.701  \\
                        & Ent\_Type &  0.737 &     0.721 &  0.762  \\ 
                        & Partial   &  0.713 &     0.700 &  0.736  \\
                        & Strict    &  0.675 &     0.663 &  0.695  \\
\midrule
\multirow{4}{*}{ReGen (Ours) T2G.CE }
                        & Exact       & \textbf{0.723}  &        \textbf{0.714} &      \textbf{0.738}  \\
                        & Ent\_Type   & \textbf{0.807}  &        \textbf{0.791} &      \textbf{0.835}  \\
                        & Partial     & \textbf{0.767}  &        \textbf{0.755} &      \textbf{0.788}  \\
                        & Strict      & \textbf{0.720}  &        \textbf{0.713} &      \textbf{0.735}  \\
\midrule
\multirow{4}{*}{ReGen (Ours) T2G.RL }
                        & Exact     &   0.720 &  0.712 &       0.734 \\
                        & Ent\_Type &   0.804 &  0.789 &       0.829 \\
                        & Partial   &   0.764 &  0.752 &       0.784 \\
                        & Strict    &   0.717 &  0.709 &       0.731 \\
\bottomrule
\end{tabular}
\caption{T2G Best results on WebNLG+ 2020 (v3.0) dataset. The top two teams were the first and second place winner of the Challeneg. Our T2G.CE model improves upon all metrics for all matching schemes, providing a new state-of-the-art results for this Challenge task. T2G.RL models, while still better than previous best results, does not improve upon its CE counterpart.}
\label{tab:web_t2g_post}
\end{table*}

\noindent\textbf{Metrics} WebNLG+ 2020 provides automatic metrics to evaluate models. For G2T, we used
BLEU, BLEU\_NLTK, METEOR, and chrF++ that are provided by the challenge. 
For T2G, F1, Precision, and Recall scores are utilized and computed for 4 levels of match: Exact, Ent\_Type, Partial and Strict as described in \citet{castro-ferreira-etal-2020-2020}, which loosely correspond to different levels of relaxation of how close a match of an entity must be to the ground truth in content and position in a triple.
Note that when generating graphs/RDFs, scoring metrics explore all possible permutations of a graph edges. For \tekgen, we use the same metrics as for WebNLG+ 2020.

\begin{table*}[ht!]
\centering
\begin{tabular}{lccccc}
%\toprule
\tekgen G2T                &  & BLEU$\uparrow$ & \multirow{2}{2cm}{\centering BLEU$\uparrow$ NLTK} & METEOR$\uparrow$   & chrF++$\uparrow$  \\
Model         &   &      &      &           &         \\
\midrule
\multirow{2}{*}{ReGen-CE}    & Val               &  0.240 & 0.241 &  0.231 &  0.400 \\
                             & Test              &  0.241 & 0.242 &  0.233 &  0.405 \\
\midrule
\multirow{2}{*}{ReGen-SCST}  & Val               &  0.258 & 0.259 &  0.240 &  0.418 \\ 
                             & Test              &  \textbf{0.262} & \textbf{0.262} &  \textbf{0.242} &  \textbf{0.422} \\ 
%\bottomrule
\end{tabular}
\caption{G2T Results for \tekgen dataset. ReGen-CE establishes a baseline on this dataset. ReGen-SCST consistently improve on the baseline on all metrics, for validation and test sets.}
\label{tab:tekgen_g2t}
\end{table*}

\section{Results}

For all  experiments, PLMs were first exposed to the target datasets (WebNLG+, \tekgen) by fine-tuning using $\Lce$ loss. They were then switched to RL training by optimizing the $\Lscst$ loss.
Although no exact recipe has been established for Seq2Seq RL-training, 
starting from a good CE model helps RL training performance in practice \citep{ranzato2016sequence, rennie2017self}. Therefore, we followed the subsequent simple approach:
During fine-tuning, the evaluations are conducted on the validation set. 
From the CE phase, the best performing model iteration is selected based on the METEOR and F1 score for the G2T and T2G tasks, respectively, to pursue RL fine-tuning. In case of G2T, potential ties in METEOR scores among candidate models, are resolved by using BLEU\_NLTK, followed by the chrF++ metric.
Note that early stopping selection of CE models led to good performance for t5-base models as well.
During the SCST phase, the best model iteration on the validation set is selected and its performance numbers on the test set are reported in our tables.

\noindent\textbf{WebNLG+ 2020 G2T}
For the WebNLG+ 2020 Challenge, the results of the top four systems for RDF-to-text task can be found in \Tab{tab:web_g2t_post} for all categories (results for seen and unseen categories are given in \Tab{tab:webnlg_cats} in the Appendix), while descriptions the top teams' systems were given in \Sec{sec:related}.
We report our G2T results for both t5-large and t5-base models as well. 
For t5-large, ReGen G2T.CE is the best model from CE fine-tuning. 
ReGen G2T.RL is best model performance for SCST training while using METEOR as reward when starting from G2T.CE model. 
\Tab{tab:web_g2t_post} shows that our CE model is better than models from all top teams, and the SCST results further improve significantly in all metrics achieving state-of-the-art results to our knowledge.
The gain obtained by SCST alone is quite significant and demonstrates the benefits of RL fine-tuning for this task.
While we report our best model results in \Tab{tab:web_g2t_post}, we also report mean and standard deviation results for multiple random number generator seeds in \Tab{tab:web_g2t_post_meanstd} in Appendix. Even in averaging results of various seeded models, we see a sustained gain from SCST for CE models in all metrics.

Multiple reward candidates were investigated (BLEU, BLEU\_NLTK, METEOR, chrF) as well as some linear combinations of pairs of them, as can be seen in \Tab{tab:abl_metrics} in Appendix. 
In \Tab{tab:abl_metrics}, for t5-large, METEOR is consistently the best SCST reward, and improves all the other metrics scores as well.
However, for `smaller' models such as t5-base, BLEU\_NLTK is revealed to be the best reward for improving BLEU performance as expected. 
Again, SCST brings significant gains across all the metrics in that case.
Note that for t5-base model, selecting a METEOR reward improves METEOR results significantly as reported in \Tab{tab:web_g2t_post_all} in Appendix. 

Another interesting fact is that early stopping of CE model G2T.CE.ES (at 5 epochs) leads to the best SCST model G2T.RL.ES for t5-base, while selecting the best CE model G2T.CE.best (at 11 epochs) still showed some gains from SCST model G2T.RL.best.
SCST needs a good starting point, but a better CE model that has seen a lot more epochs of our dataset maybe harder for SCST to stir in a better solution in the parameter space. 
Moreover, the test split contains unseen categories not present in the validation dataset which render choices based on validation sub-optimal for the test dataset.
The best models we report in this work are specialized models $\mG$. Early in our investigation, hybrid models were the best performing model for G2T reaching 0.547 BLEU, 0.543 BLEU\_NLTK and 0.417 METEOR, and first to beat the Challenge winning team. However, when batch size became larger (20-24 samples), the specialized models took the lead and retain it still.

For training, we optimized all our models using AdamW \citep{adamw-2017}, variant of the Adam optimizer with default values of $\beta=[0.9, 0.999]$ and weight decay of $10^{-2}$. For learning rate, we used $5.10^{-6}$ for all our experiments as it was better than $10^{-5}$ and $10^{-6}$ as seen in \Tab{tab:abl_lr} in Appendix. 
All our models were trained with 20-24 minibatch size on WebNLG. Further details on our experimental setup are provided in the Appendix in \Sec{sec:train}.

\noindent\textbf{WebNLG+ 2020 T2G}
Results for the Text-to-RDF task are reported in \Tab{tab:web_t2g_post} for all categories. 
Results for our best model on seen and unseen categories are given in \Tab{tab:web_t2g_cats} in Appendix.
AmazonAI and bt5 are the top performing teams. 
Again, the proposed ReGen T2G.CE model shows strong results that are better in term of  \emph{all} metrics, for \emph{all} matching categories. In themselves, these numbers are a de-facto new state-of-the-art for this dataset, as far as we know. 
SCST model T2G.RL fails to improve on this model though. 
The \emph{exact F1} metric was used as reward, but the model could never pull ahead of the CE model in our experiments.
The exact F1 metric may not be a strong enough reward to really capture the dynamics of graph generation properly for WebNLG+ as it is very rigid in its measure (one must have an exact match), although the same reward gave good results on our second dataset \tekgen. 
A more sensitive metric could possibly help. 
We even tried to use n-gram based metrics (like BLEU) but to no avail.
We further address this issue at the end on this Section.

\noindent\textbf{\tekgen G2T}
For the \tekgen dataset, we present our results on Graph-to-Text generation in \Tab{tab:tekgen_g2t}. Similar to the experiments in WebNLG+, we pick the best model during the CE fine-tuning based on the METEOR score and proceed with the RL fine-tuning. We observe that the RL fine-tuning step helps boost the test split scores on all metrics. It is worth noting that the scores are slightly underestimating the potential of our system because of the nature of the sentences in the \tekgen dataset. Unlike WebNLG+, in a paired text-graph sample in \tekgen, the linearized graph does not usually cover all the concepts described in the corresponding text. This leads to underestimating when the hypothesis is scored against the reference using n-gram metrics.

\noindent\textbf{\tekgen T2G}
Results for the Text-to-Graph for \tekgen are reported in \Tab{tab:tekgen_t2g}. Once the CE fine-tuning is done, we continue with the RL fine-tuning using exact F1 as reward. The performance is consistent with what we observe in G2T task for \tekgen, where SCST step boosts the performance of the model. Since, we reformulate this dataset (refer Section \ref{subsec:datasets}) to offer as T2G and G2T tasks, our approach is the first attempt in understanding the nature of \tekgen dataset and our methods provide a baseline for future research.
Please note that for both T2G and G2T tasks in \tekgen, we only start a t5-large PLM.

\begin{table}[ht!]
\centering
\begin{tabular}{llccc}
%\toprule
T2G           &  & F1$\uparrow$ & P$\uparrow$ & R$\uparrow$  \\
Model         &  &      &           &         \\
\midrule
\multirow{2}{*}{ReGen-CE}  & Val                 &  0.622 & 0.608 &  0.647 \\
                           & Test                &  0.619 & 0.605 &  0.643 \\
\midrule
\multirow{2}{*}{ReGen-SCST}  & Val              &  0.615 & 0.600 &  0.640 \\ 
                             & Test             &  \textbf{0.623} & \textbf{0.610} &  \textbf{0.647} \\ 
\bottomrule
\end{tabular}
\caption{T2G \tekgen Results: ReGen-CE establishes a baseline of the dataset. ReGen-SCST improves results on the test set compared to ReGen-CE.}
\label{tab:tekgen_t2g}
\end{table}

\noindent\textbf{Summary}
Results on WebNLG+ 2020 and \tekgen demonstrated that RL fine-tuning of models leads to significant improvements of results for T2G and G2T, establishing new state-of-the-art results for both tasks. For WebNLG+, T2G was a challenging task for RL fine-tuning. In further work, we plan to address this issue by investigating two points: First, look into a more sensible graph-dependent sampling for graph structures, rather than the current multinomial sampling of the best tokens at each generation step. Second, try a different reward schemes where the reward is more attuned to the challenges of graph generation as well as graph structure, allowing for some curriculum learning, or increasing the harshness of rewards gradually during training.
Results on \tekgen showed that RL fine-tuning is a viable option even on large-scale datasets.
To enrich this quantitative study of ReGen, we provide a few qualitative cherry picked results in \Tab{tab:cherry_t2g} and \Tab{tab:cherry_g2t} in Appendix.

%% file: conclusions.tex
\section{Conclusions}

In this paper, we proposed to use RL for improving upon current generation for text-to-graph and graph-to-text tasks for the WebNLG+ 2020 Challenge dataset using pre-trained LMs. We not only defined a novel Seq2Seq training of models in T2G and G2T generation tasks, but we established state-of-the-art results for WebNLG+ for both tasks, significantly improving on the previously published results. We provided extensive analyses of our results and of the steps taken to reach these improvements.
We then expanded our approach to large scale training by means of \tekgen where we demonstrated that RL fine-tuning provides a robust way to improve upon regular model fine-tuning within  a dataset that is orders of magnitude larger than the WebNLG+ starting point. We established gains despite a weaker content  overlap in text-graph data pairs for \tekgen. Along the way, we constructed subject, and relation-object boundaries from \tekgen sentence-triples pairs that we plan on releasing to benefit the research community. 

Future work will focus on developing a variant of SCST that leverages the unique structure of graph by either performing of more sensible graph-dependent sampling, or by investigating different reward schemes more attuned to integrating the content and structure of graphs.

%% file: suppl.tex
\section{Training Setup}
\label{sec:train}
All our experiments were run using NVIDIA V100 GPUs for training and validation, some  trainings were done on A100. We distributed our training to 2-4 GPUs depending on availability. Each training epoch for CE ranged from 30 minutes to 1 hour depending on number of GPUs utilized. 

Validation and testing (1,779 and 2,155 samples for testA and testB of WebNLG+ 2020) lasted from 40 minutes to 1 hour depending on machines. Computation was dominated by beam search generation as we used beam search with beam size of 5 and a max sequence length of 192 (since linearized graph sequence can be quite long). We used the official scoring scripts released by WebNLG+ 2020 Challenge to score all our experiments. The evaluation of graph being the most computationally expensive as all possible matching combinations are tested in what looks like a factorial complexity, taking scoring of set of triples larger than 8 from impractical to not feasible.

All our models were built using PyTorch. Total effective batch sizes were set to either 20 or 24 samples for our distributed training. We adjusted the batch size on each worker to ensure consistent global batch size of 20 or 24.

We did some search on learning rates for t5-large training and SCST rewards, see discussion and results in \Sec{sec:abl}.

All our trainings have a seeded random number generator for reproducibility.
We also report results on WebNLG+ 2020 G2T tasks for \emph{each} training setup by showing results for 3 models from different seeds, and provide means and standard deviations of these results in \Tab{tab:web_g2t_post_meanstd}.

\section{WebNLG+ 2020 Results per Categories for Best G2T and T2G Models}
\label{sec:webnlg_cats}

In \Tab{tab:webnlg_cats}, we are reporting results for all WebNLG+ 2020 categories for our best CE and RL models. While results for unseen categories are much worse than for seen categories,  RL fine-tuning manages to improve on both seen and unseen categories. 

\begin{table*}[ht!]
\centering
\begin{tabular}{lrcccccc}
%\toprule
WebNLG G2T Best Models    & Category & BLEU$\uparrow$ & \multirow{2}{2cm}{\centering BLEU$\uparrow$ NLTK} & METEOR$\uparrow$ & chrF++$\uparrow$  \\
                   &  &    &  &           &         \\
\midrule
Ours t5-large ReGen-CE 
& unseen  & 48.76 &     0.489  &   0.397  &   0.653 \\
& seen    & 59.73 &     0.592  &   0.433  & 0.722 \\
& all     & 55.26 &     0.549  &   0.418  &  0.694 \\
\midrule

Ours t5-large ReGen-SCST
& unseen    & 49.06 &  0.493  &  0.404  &  0.665 \\
& seen  & 61.22 &  0.605  &  0.440  &  0.734 \\
& all   & 56.25 &  0.559  &  0.425  &  0.706 \\
\bottomrule
\end{tabular}
\caption{G2T: Results for seen, unseen, and all categories subsets in WebNLG+ 2020 Challenge Test dataset. As expected, unseen categories much worse results than for seen categories. RL fine-tuning manages to improve on both seen and unseen categories.}
\label{tab:webnlg_cats}
\end{table*}

\Tab{tab:web_t2g_cats} provides results for seen, unseen and all categories for our best CE model ReGen T2G.CE
which established state-of-the-art results on T2G task of WebNLG+ 2020 Challenge dataset.

\begin{table*}[ht!]
\centering
\begin{tabular}{llccc}
%\toprule
WebNLG T2G               & Match & F1$\uparrow$ & Precision$\uparrow$ & Recall$\uparrow$  \\
ReGen  T2G.CE            &       &              &                      &        \\
\midrule
\multirow{4}{*}{unseen}  
                        & Exact     &  0.5809   &       0.5662  &     0.6069 \\
                        & Ent\_Type &  0.7014   &       0.6741  &     0.7497 \\
                        & Partial   &  0.6453   &       0.6241  &     0.6826 \\
                        & Strict    &  0.5754   &       0.5608  &     0.6012 \\
\midrule
\multirow{4}{*}{seen}
                        & Exact     &  0.8322   &       0.8286  &     0.8384  \\
                        & Ent\_Type &  0.8878   &       0.8811  &     0.8998  \\ 
                        & Partial   &  0.8604   &       0.8553  &     0.8696  \\
                        & Strict    &  0.8317   &       0.8282  &     0.8379  \\
\midrule
\multirow{4}{*}{all}
                        & Exact       &  0.7229  &        0.7144   &    0.7376 \\
                        & Ent\_Type   &  0.8067  &        0.7910   &    0.8345 \\
                        & Partial     &  0.7668  &        0.7547   &    0.7882 \\
                        & Strict      &  0.7202  &        0.7118   &    0.7349 \\
\bottomrule
\end{tabular}
\caption{T2G: Results for seen, unseen, and all categories subsets in WebNLG+ 2020 Challenge Test dataset. As expected the performance drops significantly for unseen categories and are the best for seen categories.}
\label{tab:web_t2g_cats}
\end{table*}

\section{Ablation Studies}
\label{sec:abl}
In Tables \ref{tab:abl_metrics} and \ref{tab:abl_lr} we present ablation studies of different optimized metrics and learning rates for SCST training. As can be seen from Table \ref{tab:abl_metrics}, when METEOR is used as a reward, we get the best performance across all the metrics. We also tried using a combination of multiple rewards with different scaling but did not get any gain over the single metric rewards. In Table \ref{tab:abl_lr}. we also show the effect of learning rate on SCST performance. Using $lr = 5\cdot10^{-6}$  gave us the best performance, while higher rates, such as $10^{-4}$, led to unstable training and collapse of SCST.  

\begin{table*}[ht!]
\centering
\begin{tabular}{lcccccc}
%\toprule
SCST Reward             & BLEU$\uparrow$ & \multirow{2}{2cm}{\centering BLEU$\uparrow$ NLTK} & METEOR$\uparrow $   & chrF++$\uparrow$  \\
                        &      &  &           &         \\
\midrule
BLEU         & 0.556 & 0.552 & 0.420 & 0.698 \\
BLEU NLTK    & 0.558 & 0.554 &  0.422 &  0.700\\
METEOR       & \textbf{0.563} & \textbf{0.559} & \textbf{0.425} & \textbf{0.706} \\
chrF++       & 0.554 & 0.551 & 0.423 & 0.701 \\ 
$\sfrac{1}{2}\cdot$METEOR+$\sfrac{1}{2}\cdot$BLEU NLTK                     & 0.555 & 0.551 & 0.421 & 0.699 \\
$\sfrac{2}{3}\cdot$METEOR+$\sfrac{1}{3}\cdot$BLEU NLTK                     & 0.547 & 0.543 & 0.419 & 0.697 \\
\bottomrule
\end{tabular}
\caption{Ablation study of metrics used as rewards in SCST for t5-large models. The results shown are on the test split.}
\label{tab:abl_metrics}
\end{table*}

\begin{table*}[ht!]
\centering
\begin{tabular}{rcccccc}
%\toprule
Learning Rate               & BLEU$\uparrow$ & \multirow{2}{2cm}{\centering BLEU$\uparrow$ NLTK} & METEOR$\uparrow$    & chrF++$\uparrow$  \\
                        &      &  &           &         \\
\midrule
$10^{-6}$          & 0.553 & 0.549 & 0.420 & 0.698 \\
$5\cdot10^{-6}$    & \textbf{0.558} & \textbf{0.554} &  \textbf{0.422} &  \textbf{0.700} \\
$10^{-5}$          & 0.544 & 0.542 & 0.419 & 0.696 \\
\bottomrule
\end{tabular}
\caption{Ablation study on learning rates in SCST (using BLEU NLTK as the optimized metric)}
\label{tab:abl_lr}
\end{table*}

\section{G2T Results t5-base models for SCST with METEOR Reward}

Results for SCST fine-tuning of  t5-base models using a METEOR reward are compiled in \Tab{tab:web_g2t_post_all}.
Clearly, these models achieve better METEOR results as expected since they are RL optimzed on this metric.
\begin{table*}[ht!]
\centering
\begin{tabular}{lcccccc}
%\toprule
WebNLG G2T                & BLEU$\uparrow$ & \multirow{2}{2cm}{\centering BLEU$\uparrow$ NLTK} & METEOR$\uparrow$ & chrF++$\uparrow$  \\
Team/model                    &      &      &           &         \\
\midrule
ReGen G2T.RL.ES.meteor t5-base (early CE)     &   0.527 & 0.523 &  0.413 &  0.689 \\              
ReGen G2T.RL.best.meteor t5-base (best CE)    &  0.528  & 0.526 &  0.412  &  0.681 \\
\bottomrule
\end{tabular}
\caption{G2T: Best results for t5-base fine-tuned with SCST using METEOR as reward.}
\label{tab:web_g2t_post_all}
\end{table*}

\section{G2T Results for Models from Multiple Random Seeds}

All our training have a seeded random number generator for reproducibility.
We also report the mean and standard deviations for all our G2T models. Each model setup was run 3 times using three independent and distinct seeds, following the same exact process. This is to ensure that our results are not just the product of a lucky system configuration or otherwise advantageous random shuffling of our training dataset.
The gain reported between CE and RL for our t5-large models are clearly still showing after average of all 3 models from distinct random seeds. For t5-base, gains between CE and RL are still present, albeit smaller than for our best systems.

\begin{table*}[ht!]
\centering
\begin{tabular}{lcccccc}
%\toprule
Team Name               & BLEU$\uparrow$ & \multirow{2}{2cm}{\centering BLEU$\uparrow$ NLTK} & METEOR$\uparrow$ & chrF++$\uparrow$  \\
                        &      &  &           &         \\
\midrule
ReGen G2T.CE t5-large   &  0.543$\pm$0.007 & 0.540$\pm$0.007 &  0.416$\pm$0.002 &  0.691$\pm$0.002 \\       
ReGen G2T.RL t5-large   &  0.553$\pm$0.007  & 0.550$\pm$0.007 &  0.422$\pm$0.002 &  0.702$\pm$0.003  \\
\midrule 
ReGen G2T.CE.ES t5-base (early CE) & 0.521$\pm$0.004 & 0.517$\pm$0.004 & 0.404$\pm$0.001 & 0.675$\pm$0.002 \\
ReGen G2T.RL.ES t5-base (early CE) & 0.528$\pm$0.007 & 0.523$\pm$0.007 & 0.408$\pm$0.002 & 0.682$\pm$0.003 \\
\midrule
ReGen G2T.CE.best t5-base (best CE) & 0.524$\pm$0.000 & 0.520$\pm$0.001 & 0.404$\pm$0.000 & 0.670$\pm$0.000 \\
ReGen G2T.RL.best t5-base (best CE) & 0.525$\pm$0.007 & 0.522$\pm$0.007 & 0.407$\pm$0.002 & 0.681$\pm$0.003 \\
\midrule
ReGen G2T.RL.ES.meteor t5-base (early CE) & 0.525$\pm$0.007 & 0.521$\pm$0.007 & 0.412$\pm$0.002 & 0.687$\pm$0.003 \\
\midrule
ReGen G2T.RL.best.meteor t5-base (best CE) & 0.527$\pm$0.007 & 0.524$\pm$0.007 & 0.410$\pm$0.002 & 0.686$\pm$0.003 \\
\bottomrule
\end{tabular}
\caption{Results means and standard deviations (SD), shown as mean$\pm$SD, for CE and SCST trained models (including our best results model) for a total of 3 different random number generator seeds used in training.}
\label{tab:web_g2t_post_meanstd}
\end{table*}

\section{Processed \tekgen Dataset}
\begin{figure*}[ht!]
\includegraphics[width=\textwidth]{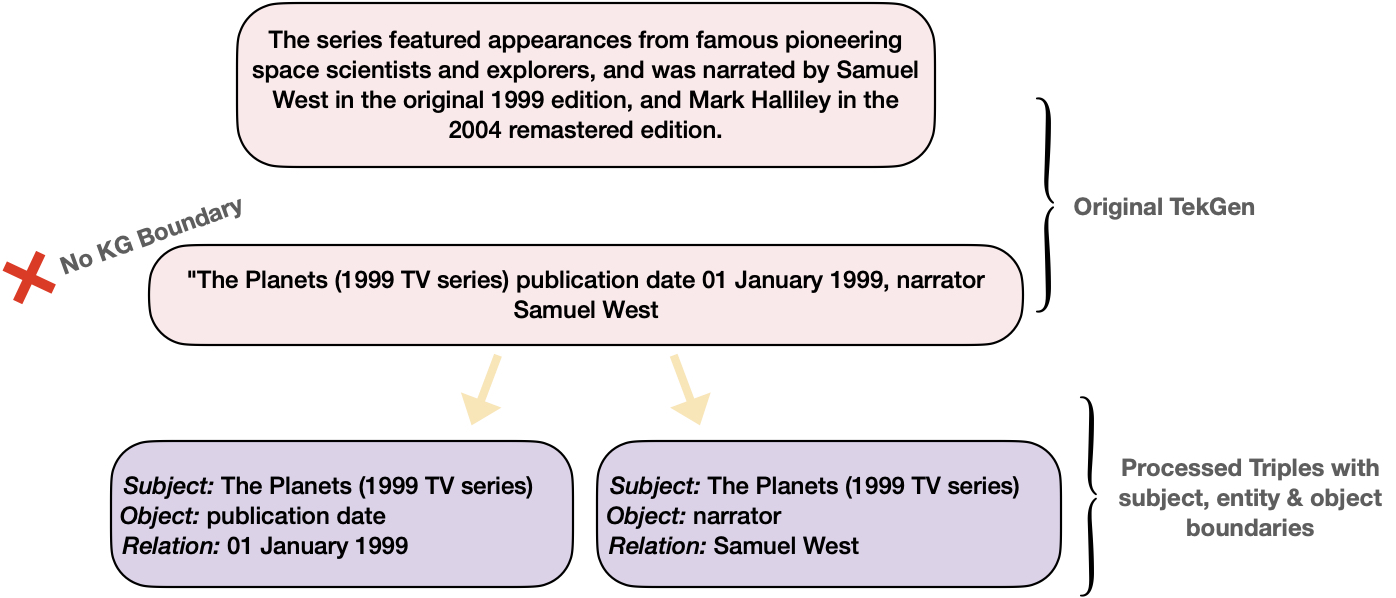}
\caption{An example from the processed \tekgen dataset. The original dataset lacks KG boundaries, which makes it difficult to evaluate T2G systems efficiently.}
\label{fig:tekgen_example}
\end{figure*}

In \Fig{fig:tekgen_example} we show an example of our processing of \tekgen dataset in establishing subject, relation, object boundaries. This enables both training and evaluating systems for T2G and G2T tasks.

\begin{table*}[ht!]
\footnotesize
\centering
\begin{tabular}{p{2cm}|p{13cm}}
%\toprule
\\
Type & Sentence / Graph                        \\
\midrule
\texttt{Source} & The Pontiac Rageous began and ended its production in 1997 on an assembly line in Detroit, a city in Michigan.\\
\texttt{Gold} & Pontiac\_Rageous $\diamondsuit$ productionStartYear $\diamondsuit$ 1997 $\diamondsuit$ Pontiac\_Rageous $\diamondsuit$ assembly $\diamondsuit$ Michigan $\diamondsuit$ Pontiac\_Rageous $\diamondsuit$ assembly $\diamondsuit$ Detroit $\diamondsuit$ Pontiac\_Rageous $\diamondsuit$ productionEndYear $\diamondsuit$ 1997 $\diamondsuit$ Detroit $\diamondsuit$ type $\diamondsuit$ City\_(Michigan)\\
\texttt{Hyp-CE} & Pontiac\_Rageous $\diamondsuit$ assembly $\diamondsuit$ Detroit $\diamondsuit$ Pontiac\_Rageous $\diamondsuit$ modelYears $\diamondsuit$ 1997 $\diamondsuit$ Pontiac\_Rageous $\diamondsuit$ modelYears $\diamondsuit$ 1997 $\diamondsuit$ Detroit $\diamondsuit$ isPartOf $\diamondsuit$ Michigan\\
\texttt{Hyp-SCST} & Pontiac\_Rageous $\diamondsuit$ assembly $\diamondsuit$ Detroit $\diamondsuit$ Pontiac\_Rageous $\diamondsuit$ modelYears $\diamondsuit$ 1997 $\diamondsuit$ Pontiac\_Rageous $\diamondsuit$ assembly $\diamondsuit$ Michigan\\
\midrule
\texttt{Source} & In the United States, where Abraham A, Ribicoff was born, African Americans are one of the ethnic groups. Abraham A. Ribicoff was married to Ruth Ribicoff.\\
\texttt{Gold} & Abraham\_A.\_Ribicoff $\diamondsuit$ spouse $\diamondsuit$ "Ruth Ribicoff" $\diamondsuit$ Abraham\_A.\_Ribicoff $\diamondsuit$ birthPlace $\diamondsuit$ United\_States $\diamondsuit$ United\_States $\diamondsuit$ ethnicGroup $\diamondsuit$ African\_Americans $\diamondsuit$ Abraham\_A.\_Ribicoff $\diamondsuit$ nationality $\diamondsuit$ United\_States\\
\texttt{Hyp-CE} & Abraham\_A.\_Ribicoff $\diamondsuit$ birthPlace $\diamondsuit$ United\_States $\diamondsuit$ Abraham\_A.\_Ribicoff $\diamondsuit$ spouse $\diamondsuit$ "Ruth Ribicoff" $\diamondsuit$ United\_States $\diamondsuit$ ethnicGroup $\diamondsuit$ African\_Americans\\
\texttt{Hyp-SCST} & Abraham\_A.\_Ribicoff $\diamondsuit$ birthPlace $\diamondsuit$ United\_States $\diamondsuit$ Abraham\_A.\_Ribicoff $\diamondsuit$ spouse $\diamondsuit$ "Ruth Ribicoff" $\diamondsuit$ Abraham\_A.\_Ribicoff $\diamondsuit$ nationality $\diamondsuit$ American $\diamondsuit$ United\_States $\diamondsuit$ ethnicGroup $\diamondsuit$ African\_Americans \\
\midrule
\texttt{Source} & Super Capers, edited by Stacy Katzman, is a 98 minute film starring Michael Rooker and Tom Sizemore.\\
\texttt{Gold} & Super\_Capers $\diamondsuit$ editing $\diamondsuit$ Stacy\_Katzman $\diamondsuit$ Super\_Capers $\diamondsuit$ starring $\diamondsuit$ Michael\_Rooker $\diamondsuit$ Super\_Capers $\diamondsuit$ starring $\diamondsuit$ Tom\_Sizemore $\diamondsuit$ Super\_Capers $\diamondsuit$ runtime | 98.0\\
\texttt{Hyp-CE} & Super\_Capers $\diamondsuit$ starring $\diamondsuit$ Tom\_Sizemore $\diamondsuit$ Super\_Capers $\diamondsuit$ timeOut $\diamondsuit$ "980.0"(minutes) $\diamondsuit$ Super\_Capers $\diamondsuit$ starring $\diamondsuit$ Michael\_Rooker $\diamondsuit$ Super\_Capers $\diamondsuit$ editor $\diamondsuit$ Stacy\_Katzman\\
\texttt{Hyp-SCST} & Super\_Capers $\diamondsuit$ starring $\diamondsuit$ Tom\_Sizemore $\diamondsuit$ Super\_Capers $\diamondsuit$ length $\diamondsuit$ 98.0 (minutes) $\diamondsuit$ Super\_Capers $\diamondsuit$ starring $\diamondsuit$ Michael\_Rooker $\diamondsuit$ Super\_Capers $\diamondsuit$ editor $\diamondsuit$ Stacy\_Katzman\\
\midrule
\texttt{Source} &  Doctor George Cary (1611-1680), Professor of Sacred Theology, lord of the manor of Clovelly, Devon, was Dean of Exeter between 1663 and 1680 (amongst other duties responsible for the maintenance and decoration of Exeter Cathedral). \\ 
\texttt{Gold} & George Cary (1611-1680) $\diamondsuit$ position held $\diamondsuit$ Dean of Exeter $\diamondsuit$ start time $\diamondsuit$ 01 January 1663 $\diamondsuit$ date of birth $\diamondsuit$ 00 1611 $\diamondsuit$ date of death $\diamondsuit$ 00 1680 \\
\texttt{Hyp-CE} & George Cary (priest) $\diamondsuit$ date of birth $\diamondsuit$ 01 January 1611 $\diamondsuit$ date of death $\diamondsuit$ 01 January 1680\\
\texttt{Hyp-SCST} & George Cary (priest) $\diamondsuit$ position held $\diamondsuit$ Dean of Exeter $\diamondsuit$ date of birth $\diamondsuit$ 01 January 1611 $\diamondsuit$ date of death $\diamondsuit$ 01 January 1680\\
\midrule
\texttt{Source} & Early general elections were held in the Bahamas on 10 April 1968. \\
\texttt{Gold} & 1968 Bahamian general election $\diamondsuit$ point in time $\diamondsuit$ 10 April 1968 $\diamondsuit$ country $\diamondsuit$ The Bahamas $\diamondsuit$ applies to jurisdiction $\diamondsuit$ The Bahamas\\
\texttt{Hyp-CE} & 1968 Bahamian general election $\diamondsuit$ point in time $\diamondsuit$ 10 April 1968\\
\texttt{Hyp-SCST} & 1968 Bahamian general election $\diamondsuit$ point in time $\diamondsuit$ 10 April 1968 $\diamondsuit$ country $\diamondsuit$ The Bahamas\\
\midrule
\texttt{Source} & The school was established on 6 January 1930, by former education minister, CWW Kannangara, who additionally founded two other colleges located in central Ceylon.\\ 
\texttt{Gold} &  Kattankudy Central College $\diamondsuit$ instance of $\diamondsuit$ School\\
\texttt{Hyp-CE} & Government Polytechnic , Colombo $\diamondsuit$ inception $\diamondsuit$ 00 1930\\
\texttt{Hyp-SCST} & Government Polytechnic , Colombo $\diamondsuit$ inception $\diamondsuit$ 00 1930 $\diamondsuit$ instance of $\diamondsuit$ School\\
\bottomrule
\end{tabular}
\caption{Few cherry-picked generation for T2G task for WebNLG+ 2020 (top three) and \tekgen (bottom three). For each source (\texttt{Text}), we show the ground truth (\texttt{Gold}) and system generated hypothesis from the best CE (\texttt{Hyp-CE}) and SCST models (\texttt{Hyp-SCST}). Note that the set of triples in WebNLG+ takes the form $\xG = [(s^1 \diamondsuit p^1 \diamondsuit o^1),\dots,(s^K \diamondsuit p^K \diamondsuit o^K)]$, whereas the same for \tekgen is of form $\xG = [s \diamondsuit (p^1 \diamondsuit o^1),\dots,(p^K \diamondsuit o^K)]$}
\label{tab:cherry_t2g}
\end{table*}

\begin{table*}[ht!]
\footnotesize
\centering
\begin{tabular}{p{2cm}|p{13cm}}
%\toprule
\\
Type & Graph / Sentence                        \\
\midrule
\texttt{Source} & McVeagh\_of\_the\_South\_Seas  $\diamondsuit$ starring  $\diamondsuit$ Harry\_Carey\_(actor\_born\_1878)  $\diamondsuit$ McVeagh\_of\_the\_South\_Seas  $\diamondsuit$ writer  $\diamondsuit$ Harry\_Carey\_(actor\_born\_1878)\\
\texttt{Gold} & Born in 1878, Harry Carey later grew up to write and star in the movie McVeagh of the South Seas. Harry Carey, born in 1878, wrote and appeared in the movie McVeagh of the South Seas. Harry Carey, who was born in 1878, wrote and starred the film of McVeagh of the South Seas.\\
\texttt{Hyp-CE} & McVeagh of the South Seas was written by Harry Carey, who was born in 1878.\\
\texttt{Hyp-SCST} & McVeagh of the South Seas was written by Harry Carey and starred the actor Harry Carey who was born in 1878.\\
\midrule
\texttt{Source} & Aleksandr\_Prudnikov  $\diamondsuit$ height  $\diamondsuit$ 185.0 (centimetres)  $\diamondsuit$ Aleksandr\_Prudnikov  $\diamondsuit$ youthclub  $\diamondsuit$ FC\_Spartak\_Moscow  $\diamondsuit$ FC\_Spartak\_Moscow  $\diamondsuit$ ground  $\diamondsuit$ Otkrytiye\_Arena\\
\texttt{Gold} & Aleksandr Prudnikov, 185cm tall played for FC Spartak Moscow's youth team. FC Spartak Moscow is based in the Otkrytiye Arena. Aleksandr Prudnikov who is 185 cm tall is a member of the youth side of FC Spartak Moscow. The home ground of FC Spartak Moscow is Otkrytiye Arena. Aleksandr Prudnikov is 185.0 cm tall and played for the FC Spartak Moscow at the Otkrytiye Arena.\\
\texttt{Hyp-CE} & Aleksandr Prudnikov is 185 cm tall and played for FC Spartak Moscow's youth team at the Otkrytiye Arena.\\
\texttt{Hyp-SCST} & Aleksandr Prudnikov is 185 cm tall and played for the youth team of FC Spartak Moscow whose home ground is the Otkrytiye Arena.\\
\midrule
\texttt{Source} & Baku\_Turkish\_Martyrs'\_Memorial  $\diamondsuit$ nativeName  $\diamondsuit$ "Türk Şehitleri Anıtı"  $\diamondsuit$ Baku\_Turkish\_Martyrs'\_Memorial  $\diamondsuit$ location  $\diamondsuit$ Azerbaijan\\
\texttt{Gold} & The Native name of the Baku Turkish Martyrs' Memorial is "Türk Şehitleri Anıtı" which is located in Azerbaijan. The native name of the Baku Turkish Martyrs' Memorial is Türk Şehitleri Anıtı located in Azerbaijan. The native name for the Baku Turkish Martyrs' Memorial is Türk Şehitleri Anıtı, which is located in Baku, Azerbaijan.\\
\texttt{Hyp-CE} & The native name of the Baku Turkish Martyrs' Memorial in Azerbaijan is Türk Şehitleri Anıtı.\\
\texttt{Hyp-SCST} & The Baku Turkish Martyrs' Memorial is located in Azerbaijan and is known locally as Türk Şehitleri Anıtı.\\
\midrule
\texttt{Source} & John Banister (anatomist)  $\diamondsuit$ occupation $\diamondsuit$ Surgeon $\diamondsuit$ date of birth $\diamondsuit$ 01 January 1533 $\diamondsuit$ date of death $\diamondsuit$ 01 January 1610\\
\texttt{Gold} & John Banister (1533-1610) was an English anatomist, surgeon and teacher.\\
\texttt{Hyp-CE} & John Banister (1533-1610) was an English surgeon.\\
\texttt{Hyp-SCST} & John Banister (1533-1610) was an English surgeon and anatomist.\\
\midrule
\texttt{Source} & WNPT (TV)  $\diamondsuit$ country $\diamondsuit$ United States $\diamondsuit$ instance of $\diamondsuit$ Television station\\
\texttt{Gold} & WNPT, virtual channel 8 (VHF digital channel 7), is a PBS member television station licensed to Nashville, Tennessee, United States.\\
\texttt{Hyp-CE} & WNPT, virtual channel 3 (UHF digital channel 15), is a Fox-affiliated television station licensed to Portland, Oregon, United States.\\
\texttt{Hyp-SCST} & WNPT, virtual channel 4 (UHF digital channel 16), is a Public Broadcasting Service (PBS) member television station licensed to Portland, Oregon, United States.\\
\midrule
\texttt{Source} & Our Lady of the Presentation Cathedral, Natal  $\diamondsuit$ inception $\diamondsuit$ 21 November 1988\\
\texttt{Gold} & Our Lady of the Presentation Cathedral, Natal was inaugurated on November 21, 1988, and is located in the district of Cidade Alta in Natal, capital of the Brazilian state of Rio Grande do Norte.\\
\texttt{Hyp-CE} & Our Lady of the Presentation Cathedral, Natal was built in 1988.\\
\texttt{Hyp-SCST} & Our Lady of the Presentation Cathedral, Natal was consecrated on 21 November 1988.\\
\midrule
\bottomrule
\end{tabular}
\caption{Few cherry-picked generation for G2T task for WebNLG+ 2020 (top three) and \tekgen (bottom three). For each source (\texttt{Graph}), we show the ground truth (\texttt{Gold}) and system generated hypothesis from the best CE (\texttt{Hyp-CE}) and SCST models (\texttt{Hyp-SCST}). Note that the set of triples in WebNLG+ 2020 takes the form $\xG = [(s^1 \diamondsuit p^1 \diamondsuit o^1),\dots,(s^K \diamondsuit p^K \diamondsuit o^K)]$, whereas the same for \tekgen is of form $\xG = [s \diamondsuit (p^1 \diamondsuit o^1),\dots,(p^K \diamondsuit o^K)]$}
\label{tab:cherry_g2t}
\end{table*}